\documentclass[conference]{IEEEtran}
\usepackage{cite}
\usepackage{amsmath,amssymb,amsfonts}
\usepackage{algorithmic}
\usepackage{graphicx}
\usepackage{subfigure}

\usepackage{hyperref}
\usepackage{fixltx2e}
\usepackage{balance}
\usepackage{textcomp}
\def\BibTeX{{\rm B\kern-.05em{\sc i\kern-.025em b}\kern-.08em
    T\kern-.1667em\lower.7ex\hbox{E}\kern-.125emX}}
    
\def\eg{\textit{e.g}. } 
\def\ie{\textit{i.e}. } 
\def\cf{\textit{c.f}. }

\begin{document}

\title{Deep Learning in the Field of \\Biometric Template Protection: An Overview}

\author{\IEEEauthorblockN{C. Rathgeb$^*$, J. Kolberg$^*$, A. Uhl$^\dagger$, C. Busch$^{*\ddagger}$}
\IEEEauthorblockA{$^*$da/sec -- Biometrics and Internet Security Research Group, 
 Hochschule Darmstadt, Germany \\
$^\dagger$WaveLab -- Multimedia Signal Processing and Security Lab, Universität Salzburg, Austria \\
$^\ddagger$Norwegian Biometrics Laboratory,  Norwegian University of Science and Technology, Norway  \\
\texttt{\{christian.rathgeb,jascha.kolberg,christoph.busch\}@h-da.de}, \texttt{uhl@cs.sbg.ac.at}}
}

\maketitle

\begin{abstract}
Today, deep learning represents the most popular and successful form of machine learning. Deep learning has revolutionised the field of pattern recognition, including biometric recognition. Biometric systems utilising deep learning have been shown to achieve auspicious recognition accuracy, surpassing human performance. Apart from said breakthrough advances in terms of biometric performance, the use of deep learning was reported to impact different covariates of biometrics such as algorithmic fairness, vulnerability to attacks, or template protection.  

Technologies of biometric template protection are designed to enable a secure and privacy-preserving deployment of biometrics. In the recent past, deep learning techniques have been frequently applied in biometric template protection systems for various purposes. This work provides an overview of how advances in deep learning take influence on the field of biometric template protection. The interrelation between improved biometric performance rates and security in biometric template protection is elaborated. Further, the use of deep learning for obtaining feature representations that are suitable for biometric template protection is discussed. Novel methods that apply deep learning to achieve various goals of biometric template protection are surveyed along with deep learning-based attacks. 
\end{abstract}


\section{Introduction}
\label{sec:introduction}
Deep
learning-based methods represent the current state-of-the-art for solving pattern recognition tasks \cite{LecBen15,GoodBengCour16}. In recent years, advances in deep learning have led to remarkable performance improvements in numerous areas of pattern recognition including biometric recognition \cite{Jain-HandbookBiometrics-2007,
Sundararajan-DeepLearningBiometrics-2018}. These developments have further facilitated the incorporation of biometric technologies in personal, commercial, and governmental identity management applications as evidenced by various market value studies \cite{Pascu-BiometricMarketValue-2020,MM-BiometricMarket-2018}.

\begin{figure}[!t]
\centering
\includegraphics[height=3.5cm]{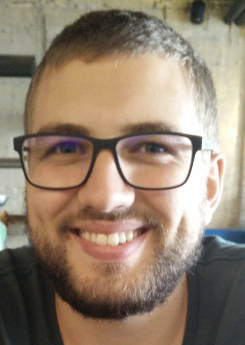} \includegraphics[height=3.5cm]{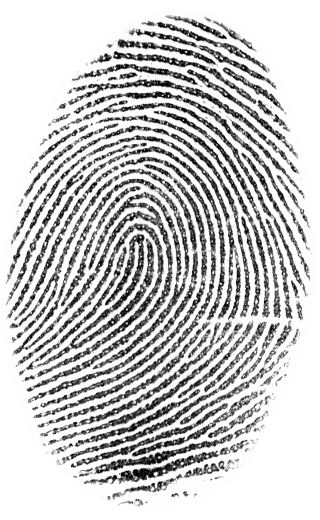} \includegraphics[height=3.5cm]{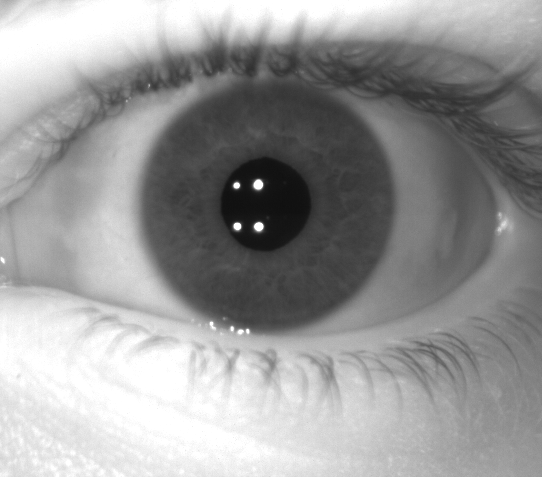}
\caption{Examples of physiological biometric characteristics. From left to right: face, fingerprint, and iris of a single subject.}
\label{fig:bio_example}\vspace{-0.4cm}
\end{figure}

In the international standard ISO/IEC 2382-37 \cite{ISO-Vocabulary-2017}, the term \emph{biometrics} is defined as: “automated recognition of individuals based on their biological and behavioural characteristics”. There exist several highly distinctive biological and behavioural characteristics which are suitable for recognising individuals, examples of prominent physiological characteristics are shown in figure~\ref{fig:bio_example}.

In an automated biometric system, a capture device (\eg a camera) is used to acquire a biometric sample (\eg facial image) during the enrolment process. Signal processing algorithms are subsequently applied to the biometric sample, which pre-process it (\eg detection and normalisation of the face), estimate the quality of the acquired sample, and extract discriminative features from it. The resulting feature vector is finally stored as reference template. At the time of authentication, another biometric sample is processed in the same way resulting in a probe template. Comparison and decision algorithms enable ascertaining of similarity of a reference and a probe template by comparing the corresponding feature vectors and establish whether or not the two biometric samples belong to the same source.


For a long period of time, biometric signal processing algorithms employed handcrafted
features (\eg texture descriptors). Nowadays, the use of deep learning has become increasingly popular. Deep learning-based biometric technologies utilise massive training datasets to learn rich and compact representations of biometric characteristics. Until now, deep learning has been applied to the vast majority of biometric characteristics including, \eg face \cite{Parkhi-VGGFace-BMVC-2015}, fingerprint \cite{Tang-FingerNet-2017}, iris \cite{LIU2016}, vein \cite{Das19}, online signatures \cite{Tolosana18}, or gait \cite{Shiraga16}. For a review of deep learning techniques applied within biometrics, the interested reader is referred to \cite{Bhanu17,Sundararajan-DeepLearningBiometrics-2018,Jiang2020}.

The success and ubiquitous use of biometric technologies has raised different privacy concerns due to the potential misuse of biometric data. In response to this issue, current privacy regulations contain provisions on the storage and use of biometric data. For instance, the General Data Protection Regulation (GDPR) \cite{EU-GDPR-2016} generally classifies biometric data as sensitive data which require protection. It is important to note that this applies to biometric samples as well as biometric templates. While the latter might not be human-interpretable, these can be misused to launch cross-matching attacks and  approximations of biometric samples can be reconstructed from unprotected biometric templates \cite{GomezBarrero-SurveyInverseBiometrics-CS-2020}. It is well-known that traditional encryption methods are unsuitable for protecting biometric data, due to the natural intra-class variance of biometric characteristics. More precisely, biometric variance prevents the usage of symmetric cryptography and traditional hash functions with biometric input data since slight changes in the unprotected domain automatically leads to drastic changes in the protected domain. Consequently, the use of conventional cryptographic methods does not enable permanent protection since it would require a decryption of protected biometric data prior to the comparison. 

\emph{Biometric template protection} \cite{Rathgeb11e,BNandakumar15a} 
refers to a class of technologies which are capable of permanently protecting biometric reference data. In contrast to conventional biometric recognition methods, biometric template protection schemes generate protected reference templates (while unprotected biometric data is discarded). Protected templates prevent from reconstruction attacks, but nevertheless make it possible to perform a biometric comparison in the protected domain. Moreover, template protection scheme typically enable the incorporation of random parameters in the generation process of protected templates. Thereby, protected templates become variable and can be changed which prevents from cross-matching attacks. The operating principle of biometric template protection is detailed in section~\ref{sec:btp}. 

\begin{figure}[!t]
\centering
\includegraphics[width=\linewidth]{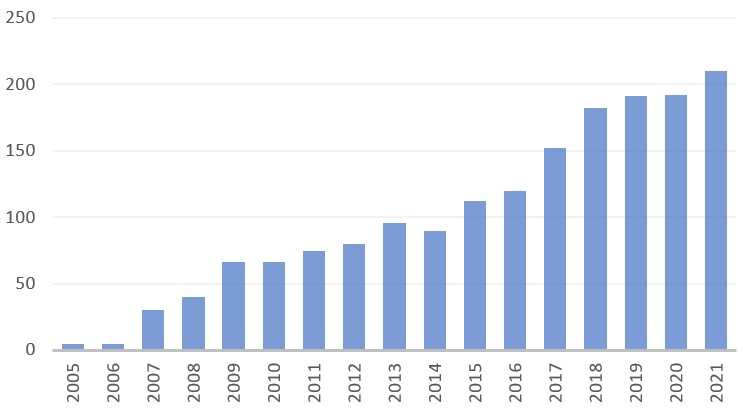}
\caption{Annual amount of scientific publications containing the terms ``biometric template protection'' or ``biometric information protection''.\textsuperscript{1}}
\footnotesize\textsuperscript{1}\url{https://app.dimensions.ai/} (encyclopaedia entries have not been counted)
\label{fig:btp_stats}\vspace{-0.3cm}
\end{figure}

Biometric template protection has been an active field of research for more than two decades. Figure~\ref{fig:btp_stats} shows the annual number of scientific publications dealing with biometric template protection. Clearly, a non-declining interest in biometric template protection is observable. It is reasonable to assume that biometric template protection approaches introduced in the last couple of years are incorporating deep learning-based methods. In many cases, the feature extraction step is based on deep learning techniques in order to achieve competitive biometric performance which in turn enhances privacy protection as well as security (see section~\ref{sec:sec}). However, beyond the use of deep learning for feature extraction, there exist numerous further processing steps in biometric template protection that can benefit from deep learning techniques, \eg feature type transformation or feature alignment. This work is intended to provide an overview of different approaches of incorporating deep learning methods into various processing steps of biometric template protection schemes with the overall goal of privacy protection. In addition, deep learning-based attacks on biometric template protection schemes are summarised. Note that in contrast to other surveys in the field, this work does not intend to completely survey existing template protection schemes. In contrast, this work is meant to capture general trends of broad interest in the application of deep learning techniques in the field of biometric template protection. It is meant to be a milepost along the research area of biometric template protection, offering guidance about future research directions.

\setcounter{footnote}{+1} 

This work is organised as follows: fundamentals of biometric template protection schemes are briefly introduced in section~\ref{sec:btp}. The impact of accuracy gains in deep learning-based biometric systems on biometric template protection are discussed in section~\ref{sec:sec}. Section~\ref{sec:ftt} surveys the use of deep learning techniques for the purpose of feature type transformation. Deep learning-based methods which are employed to achieve privacy protection are summarised in section~\ref{sec:btpdl}. Subsequently, deep learning-based attacks on biometric template protection schemes are discussed in section~\ref{sec:attacks}. Section~\ref{sec:further} lists further works which incorporate deep learning techniques that are relevant for biometric template protection. Finally, a summary is given in section~\ref{sec:summary}. 

\begin{figure*}[!t]
\centering
\subfigure[feature level]{
\includegraphics[width=0.7\linewidth]{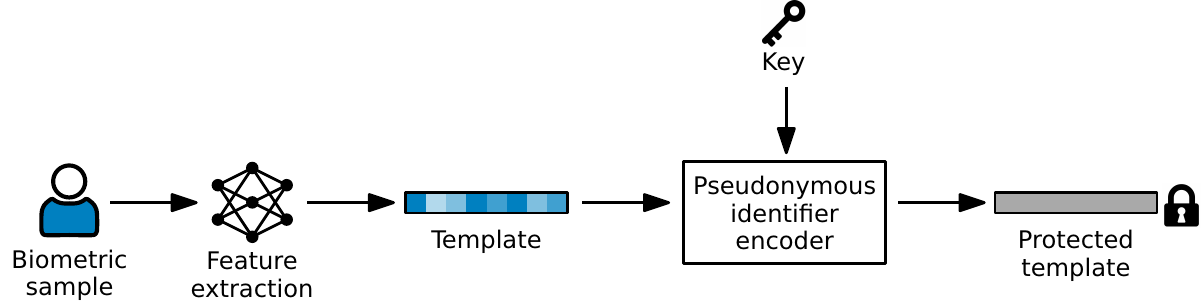}}
\subfigure[signal level]{
\includegraphics[width=0.7\linewidth]{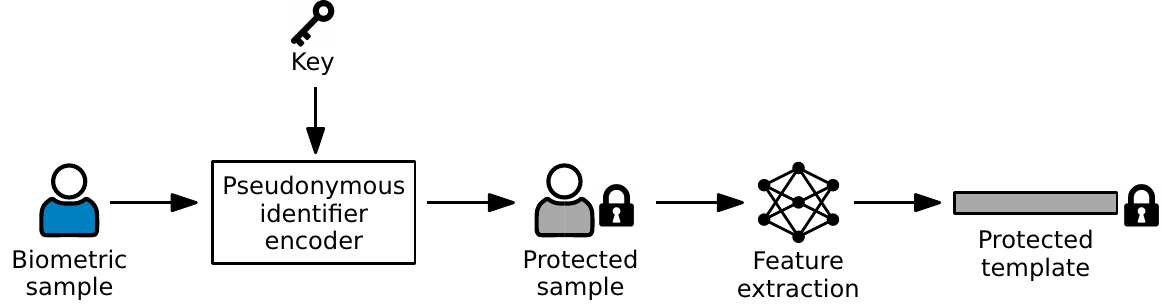}}
\caption{Pseudonymous identifier encoders applied at (a) feature level and (b) signal level.}
\label{fig:btp}\vspace{-0.3cm}
\end{figure*}

\section{Biometric Template Protection}\label{sec:btp}

For comprehensive surveys on pre-deep learning biometric template protection the reader is referred to \cite{Rathgeb11e,BNandakumar15a,Barni-BTP-SPM-2015,Sandhya-BIPsurvey-BSP-2017}. Biometric template protection schemes use auxiliary data to obtain pseudonymous identifiers from unprotected biometric data. Biometric comparisons are then performed via pseudonymous identifiers while unprotected biometric data are discarded. Biometric template protection methods are commonly categorised as \emph{cancelable biometrics}, \emph{biometric cryptosystems}, and \emph{biometrics in the encrypted domain}. Cancelable biometrics employ transforms in signal or feature domain which enable a biometric comparison in the transformed (encrypted) domain \cite{Patel-CancelableBiometrics-2015, Kumar-CancelableBiometricsSurvey-AIR-2019}. In contrast, the majority of biometric cryptosystems binds a key to a biometric feature vector resulting in a protected template. Biometric comparison is then performed indirectly by verifying the correctness of a retrieved key \cite{BUludag04a}. That is, biometric cryptosystems further allow for the derivation of digital keys from protected biometric templates, \eg fuzzy commitment \cite{BJuels99a} and fuzzy vault scheme \cite{BJuels02a}. Alternatively, \emph{homomorphic encryption} has frequently been suggested for biometric template protection \cite{Aguilar-Homomorphic-2013,Acar-HESurvey-CSUR-2018}. Homomorphic encryption makes it possible to compute operations in the encrypted domain which are functionally equivalent to those in the plaintext domain and thus enables the estimation of certain distances between protected biometric templates.

A general framework for biometric template protection methods is defined in ISO/IEC 24745 \cite{ISO11-TemplateProtection}. At enrolment, biometric data $M$ and some random secret $k$ are fed to a pseudonymous identifier encoder (PIE). The randomness $k$ can be a subject- or application-specific secret. PIE then generates a pseudonymous identifier $\mathbf{PI}$ and auxiliary data $\mathbf{AD}$, $\textrm{PIE}(M,k)=[\mathbf{PI},\mathbf{AD}]$. That is, $\mathbf{PI}$ and $\mathbf{AD}$ constitute the protected template. In detail, $\mathbf{PI}$ represents a protected identity and $\mathbf{AD}$ is subject-specific auxiliary data which might as well be protected. $\mathbf{AD}$ is used to reproduce $\mathbf{PI}$ in the authentication process. The unprotected biometric data $M$ is discarded at the end of the enrolment procedure. At the time of authentication, the pseudonymous identifier recorder (PIR) takes biometric data $M^*$ and $\mathbf{AD}$ as input and calculates another pseudonymous identifier $\mathbf{PI}^*$. This process may also require a repeated presentation of $k$, \eg in the case of cancelable biometrics. Subsequently, the pseudonymous identifier comparator (PIC) compares $\mathbf{PI}$ with the stored $\mathbf{PI}^*$, $\textrm{PIC}(\mathbf{PI},\mathbf{PI}^*)=s$. Depending on comparators, the comparison result $s$ is either a binary decision (yes/no) or a similarity score which is then compared against a threshold in order to obtain a binary decision. 

Nearly all biometric template protection schemes can be described in the aforementioned framework. It is important to note that the biometric data $M$ can be of different representations. While the majority of biometric template protection methods are applied to biometric feature vectors (templates), some are applied at signal level, \eg in the image domain. The difference in applying the PEI in feature and signal domain is illustrated in figure~\ref{fig:btp}. In such schemes, a subsequent feature extraction process is necessary which can be seen as part of the PIE.

The requirements on biometric template protection schemes are defined in ISO/IEC IS 24745 \cite{ISO11-TemplateProtection}:

\begin{itemize}
\item \textbf{Unlinkability}: the infeasibility of determining if two or more protected templates were derived from the same biometric instance, \eg face. By fulfilling this property, cross-matching across different databases is prevented. It can only be achieved by incorporating a random secret $k$ in PIE. 
\item \textbf{Irreversibility}: the infeasibility of reconstructing the original biometric data given a protected identity $\mathbf{PI}$ and its corresponding auxiliary data $\mathbf{AD}$. In certain attack models, the secret $k$ may additionally be available\footnote{In a scenario, where $k$ is known to the attacker, irreversibility can not be achieved by some template protection schemes, \eg homomorphic encryption.}. To achieve this goal the PIE needs to extract the protected template in an irreversible manner using a one-way function. With this property fulfilled, the privacy protection is enhanced, and additionally the security of the system is increased against reconstruction attacks.
\item \textbf{Renewability}: the possibility of revoking old protected templates and creating new ones from the same biometric instance and/or sample, \eg face image, without the subject needing to re-enroll. With this property fulfilled, it is possible to revoke and reissue the templates in case the database is compromised, thereby preventing misuse. Again, this goal is achieved through the incorporation of random secrets into PIE.
\item \textbf{Performance preservation}: the requirement of the biometric performance not being significantly impaired by the protection scheme.
\end{itemize}

Some research efforts and standardisation activities have been devoted to establishing metrics for evaluating the aforementioned properties of biometric template protection schemes, \eg in \cite{Nagar10a,Wang12a,BSimoens12a,Gomez-Barrero-FrameworkForUnlinkability-TemplateProtection-2018,ISO-IEC-30136}. Nonetheless, additional specific cryptanalytic methods may be necessary to precisely estimate the security/privacy protection achieved by a particular template protection scheme. Different attack models for describing scenarios and assumptions of attacks on biometric template protection schemes are standardised in ISO/IEC IS 30136 \cite{ISO-IEC-30136}. The most restrictive model is referred to as na{\"i}ve model in which an attacker has neither information of the biometric template protection scheme, nor owns a large biometric database. However, it is common practise to analysed template protection schemes under Kerckhoffs‘s principle. In this general model, an adversary is assumed to possess full knowledge of the template protection algorithm. In addition, the adversary may have access to one or more protected templates from one or more databases. In the most challenging full disclosure model, the general model is augmented by disclosing the secrets (if any) to the attacker.

\section{Accuracy and Privacy Protection}\label{sec:sec}
As mentioned before, deep learning-based pattern recognition has the edge over previous handcrafted methods \cite{Liu-TextureSurvey-2019}. Obviously, the availability of sufficient training data is a prerequisite deep learning-based approaches. In the field of biometric recognition, the use of deep learning techniques is continuously improving biometric performance, \ie recognition accuracy, for various biometric characteristics \cite{Sundararajan-DeepLearningBiometrics-2018}. Performance gains have been achieved by incorporating deep learning-based methods at various processing stages of biometric systems including detection, segmentation, quality control, or feature extraction. The latter processing step has received most attention in the development of deep learning techniques. It should be noted that said performance gains are usually more pronounced when a large amount of training data is available and vice versa.

The performance of biometric algorithms is determined by various metrics standardised in ISO/IEC IS 19795-1 \cite{ISO-IEC-19795-1-Framework-210216}. Most importantly, the following metrics are reported to measure recognition accuracy of a biometric verification system: 

\begin{itemize}
\item \textbf{False Non-Match Rate (FNMR)}: the FNMR is the proportion of completed mated comparison trials that result in a comparison decision of ``non-match''.
\item \textbf{False Match Rate (FMR)}: the FMR is the proportion of a specified set of completed non-mated comparison trials that result in a comparison decision of ``match''.
\end{itemize}

In template protection schemes, these performance metrics are also applicable. However, for some classes of schemes, \eg biometric cryptosystems, the configuration of a desired operation point may turn out to be more complex. While the FNMR is related to the usability of the biometric system, the FMR is a measure of security. Both rates depend on the decison threshold of the biometric system. There is a relation between the FMR and the security of a template protection scheme. The so-called False Accept Security (FAS) is expressed in bits and corresponds to an encryption using a key of that bit-length. The FAS provides a good approximation of the security of a template protection scheme because the false accept attack is typically one of the most efficient attacks.  In a false accept attack, the adversary iteratively simulates non-mated authentication trails until a false match is reached. The FAS (in bits) is often estimated as  $\textrm{FAS}=-\log_2(\textrm{FMR})$. For instance, a biometric system operated at a FMR of 0.01\% achieves a security level of approximately 13 bits. However, it should be noted that an attacker can deviate from the parameters corresponding to the operational FMR. Additionally, the time required for performing a single authentication attempt should be taken into account when estimating the FAS. Alternatively to the FAS, many published works report the Brute Force Security (BFS) as a measure of security. Usually, the BFS is derived from the size of an employed secret $k$ ($\widehat{=}$ key space). However, the BFS often overestimates the effective security of a biometric template protection scheme.

The aforementioned FAS directly relates to the privacy protection cappabilities of a biometric template protection scheme. Assume that an attacker requires at most $\textrm{FMR}^{-1}$ authentication attempts to find a biometric data $M'$ which results in a false match for a target protected template. This means, $M'$ is sufficiently close to $M$ that was used to generate an attacked protected template $[\mathbf{PI},\mathbf{AD}]$. Starting from $M'$, for the majority of template protection schemes, it is then possible for the attacker to reconstruct the original biometric data $M$. In case $M$ is a binary feature vector, the attacker could iteratively flip bits of $M'$ until a false non-match is reached resulting in $\hat{M}$. Hence, the Hamming distance of $\hat{M}$ and $M$ is exactly one bit above the threshold for reaching a false match. The attacker would then iteratively flip bits of $\hat{M}$. If flipping the $i$-th bit results in a false match, the bit at the $i$-th position of $\hat{M}$ is equal to the flipped bit. If not, the bit at the  $i$-th position of $\hat{M}$ has already been correct. Similar attacks can be applied for non-binary data representations.

Many template protection concepts, in particular biometric cryptosystems, have been investigated several years ago, \cf figure~\ref{fig:btp_stats}. Hence, such earlier approaches have been applied to biometric systems utilising handcrafted feature extractors. Compared to such handcrafted approaches, deep learning-based methods have been shown to improve biometric performance for systems based on different biometric characteristics \cite{Sundararajan-DeepLearningBiometrics-2018}. This means that deep learning-based biometric systems can be potentially operated at lower FMRs and, thus, provide a higher level of security. The same applies to biometric template protection schemes utilising deep learning-based techniques. For instance, recently proposed biometric cryptosystems utilizing deep learning-based face recognition can be operated at FMRs significantly below 0.1\%, \eg in \cite{RATHGEB2022102539,Gilkalaye19}, while comparable older approaches based on handcrafted feature extractors report unpractical FMRs above 10\%, \eg in \cite{Wu10a}. While many biometric performance benchmarks have been published for various biometric characteristics, a comprehensive empirical comparison between earlier handcrafted methods and current deep learning-based methods is still missing. 

Through reduced error rates the level of privacy protection is significantly increased in such systems. However, some studies have also reported an increased vulnerability of deep-learning-based biometric systems to certain attacks subjected to their enhanced generalisation capabilities, \eg in \cite{Mohammadi18,Scherhag-MorphingSurvey-2019}.

\section{Feature Type Transformations}\label{sec:ftt}

Biometric feature vectors, \ie templates, can be of different  representations depending on the type of feature elements (real, integer, or binary), their dimension and if they are fixed-length or variable-length. Common feature representations have been established for templates of different biometric characteristics, \eg minutiae sets for fingerprints or binary codes for iris. However, biometric template protection schemes, in particular biometric cryptosystems, may require templates in a distinct type of feature representation, \eg fixed-length binary strings for the fuzzy commitment scheme \cite{BJuels99a}. To make biometric templates compatible to a template protection scheme, so-called feature type transformation processes may be required \cite{Lim-BiometricFeatureTypeTransformation-IEEE-2015}. 

Deep learning-based biometric feature extraction methods are commonly trained using differentiable loss functions, \eg Euclidean distance. Therefore, extracted templates are usually represented as real-valued vector
$\mathbf{v}$ of fixed dimension $n$, $\mathbf{v} \in \mathbb{R}^n$. Compared to variable sized feature vectors, this is favourable since it is expected to yield an equal level of security across subjects. It is important to note that the extraction of fixed-length vectors is also possible for biometric characteristics where templates are usually not of variable size, \eg fingerprints. In \cite{Engelsma21}, it was shown that deep networks can learn to extract fixed-length fingerprint representations incorporating domain knowledge, such as alignment and keypoint (minutiae) detection\footnote{Note that a fixed-length fingerprint representation is different from fixed-length minutiae (vicinity) representations, \eg in \cite{Cappelli-MCC-PAMI-2010}, which usually result in a variable length fingerprint representation.}. 

\begin{figure}[!t]
\vspace{-0.0cm}
\centering
\includegraphics[width=0.5\linewidth]{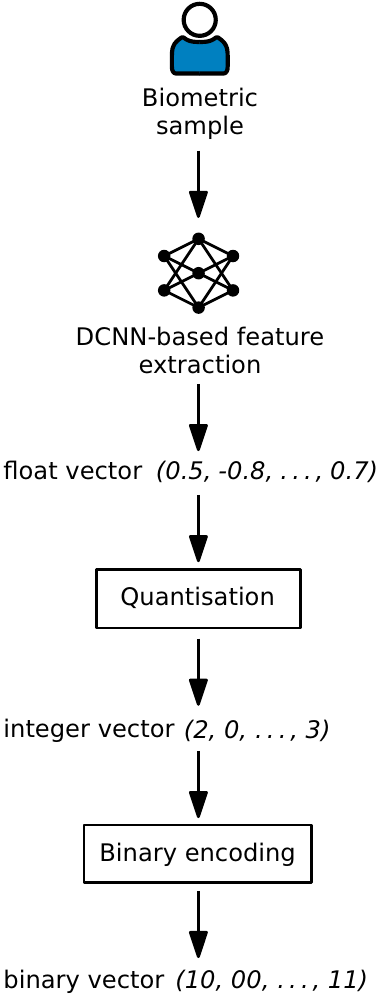}\vspace{-0.1cm}
\caption{Feature type transformation: creating integer and binary vectors from float vectors extracted by deep neural networks.}\label{fig:btp_ft}\vspace{-0.2cm}
\end{figure}

Fixed-length real-valued vectors can be mapped to other representations using common procedures \cite{Lim-BiometricFeatureTypeTransformation-IEEE-2015}. Based on statistical information about feature distributions, the feature space of each feature element can be divided in (equally probable) integer-labelled intervals. Subsequently, a quantisation can be performed by mapping each feature element to the integer value representing its interval. This results in a quantised feature vector $\mathbf{q}$ of same length, $\mathbf{q} \in \mathbb{N}_0^n$.  To obtain a binary feature vector $\mathbf{b}$, each feature element of $\mathbf{q}$ is mapped to a binary string of length $m$. Different binary encoding methods have been suggested in the scientific literature \cite{Lim-BiometricFeatureTypeTransformation-IEEE-2015}. The resulting binary vector consists consist of $nm$ bits, $\mathbf{b} \in \{0,1\}^{nm}$. These feature type  transformation steps, which are illustrated in figure \ref{fig:btp_ft}, may result in a loss of biometric performance due to information loss cause by coarse quantisation. However, if parameters are chosen appropriately biometric performance may be maintained, \eg in \cite{Drozdowski-DeepFaceBinarisation-ICIP-2018}. A binary vector $\mathbf{b}$ of length $nm$ can be transformed back to an integer vector $\mathbf{q}$ of length $n$ by mapping consecutive chunks of $m$ bits to their decimal representation. Further, $\mathbf{b}$ can be  transformed to an integer set $\mathbf{s}$ of variable length. For instance, this feature set can consist of all indexes of 1s in the binary vector,  $\mathbf{s}=\{i|b_i=1\}$ with $b_i \in \mathbf{b}$. This feature type transformation has been successfully applied to feature vectors obtained by deep learning-based feature extractors, \eg in \cite{RATHGEB2022102539}. 

Compact, \eg binary, representations (which additionally turn out to be beneficial for workload reduction in biometric identification systems \cite{Drozdowski-WorkloadSurvey-IET-2019}) can also be extracted by deep learning techniques. The term \emph{deep hashing} has been coined as an umbrella term for methods which aim at extracting compact and stable representations with deep learning techniques \cite{Cao17}. If applied to biometric data, such methods would need to overcome intra-class variations. Consequently, a reliable extraction of stable feature vectors (deep hashes) would enable a subsequent application of conventional (and provable secure) cryptographic algorithms for the purpose of template protection. In the recent past, deep hashes have been extracted from different biometric characteristics in various ways, \eg in \cite{Jindal18,Zhang20,Cui20,Talreja21}, including multi-biometrics \cite{Talreja21}. Most of the proposed schemes, however, are focusing on  facial images and have been recently surveyed in \cite{Krivokuca21}.

It is well-known that a single biometric characteristic, \eg  a single fingerprint or face, contains an insuffcient amount of effective entropy to resist attacks exploiting the feature distribution, specifically false accept attacks. Therefore, several researchers have proposed multi-biometric template protection systems \cite{Rathgeb12x}, \eg in \cite{BNagar12a,GomezBarrero-MBTPwithHE-PR-2017}. In multi-biometric template protection systems, the fusion process  should be performed at the feature level to achieve high security levels \cite{Merkle12a}. In contrast, separate storage of multiple protected biometric templates would enable parallelized attacks \cite{Rathgeb17a} (analogous to the use of multiple short passwords compared to one long password). In multi-biometric template protection schemes, feature type transformation methods become increasingly important. To conduct a feature level fusion, it is vital that features of multiple biometric sources are of the same representation. As mentioned before, this is generally feasible for the application of deep learning-based feature extraction methods. 

Note that in case a fusion of different biometric characteristics is performed, it is likely that corresponding templates are of different size. This means, in a concatenation of templates (of the same representation) an implicit weighting is introduced according to the template sizes, \ie larger templates are stronger weighted and vice versa. Such imbalance can in turn have a negative impact on the overall security of a biometric template protection system, since an attacker could merely focus on launching a false accept attack on the biometric characteristic which constitutes the largest relative part of the fused template. Further, it is important to note that templates extracted from different biometric characteristics may exhibit varying biometric variances (intra- and inter-class).

\section{Template Protection with Deep Networks}\label{sec:btpdl}

\begin{figure}[!t]
\vspace{-0.0cm}
\centering
\includegraphics[width=0.9\linewidth]{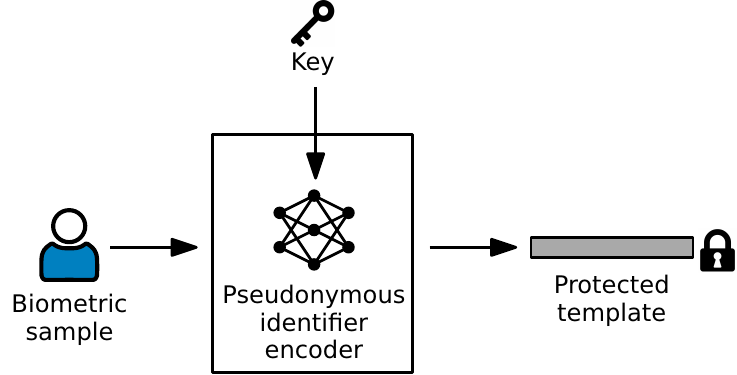}\vspace{-0.1cm}
\caption{Application of deep neural network for biometric template protection.}\label{fig:btp_dl}\vspace{-0.2cm}
\end{figure}

Very recently, first attempts have been made to directly incorporate biometric template protection into deep learning-based feature extraction. The key idea behind these approaches is to embed some randomness into neural network-based feature extraction methods. That is, the neural network itself serves as PIE taking a biometric sample $M$ and a random key $k$ as input, see figure~\ref{fig:btp_dl}. This can be achieved by introducing a key-based random activation of neurons, \ie a random sub-network selection, or random permutation, \eg in \cite{Mai21}. Such a network can be applied subsequently to an existing network trained for biometric recognition. Alternatively, networks can be trained from scratch or a pre-trained model can be adapted to achieve template protection properties \cite{Pinto21}. Moreover, researchers have suggested special loss functions, \eg in \cite{Pinto21,Lee21}, that may even incorporate a comparison of keys as suggested in \cite{Pinto21}.

The aforementioned concepts have mostly been applied to facial data \cite{Krivokuca21}. However, similar schemes have already been proposed for other biometric characteristics as well as multi-biometric systems, \eg in \cite{Kim22,Abdellatef20}. While the reported results of these very recently proposed methods are promising, a rigorous analysis of potential vulnerabilities is needed.

So-called \emph{privacy-enhancing biometrics} have been recently proposed by various research groups. These methods do not directly fall under the category of biometric template protection. In contrast to traditional template protection schemes, these methods attempt to only remove (or conceal) soft-biometric information, \eg gender or age, from biometric data (while leaving other useful information unchanged). In other words, these approaches could be seen as attempting to fulfill the requirement of irreversibility for soft biometric attributes while unlinkability or renewability are not intended properties \cite{MelziPETs}. Consequently,  privacy-enhancing biometrics do not require the incorporation of a random secret. Like cancelable biometric transformations, privacy-enhancing functions can be applied at either image level, representation level, or at inference level.  At image level deep learning-based techniques, \eg convolutional autoencoder, can be applied to perturb images with the goal of concealing soft-biometric information \cite{Mirjalili18}. At representation level it is possible to (re-)train a biometric recognition system while incorporating soft-biometric classifiers in the used loss function. Thereby, the utility of the extracted features should be maximised while at the same time soft-biometric information should be minimised.  

The majority of works on soft-biometric privacy enhancement focuses on face recognition \cite{Meden21a}, while the underlying concepts have been shown to be applicable to other biometric characteristics from which it is possible to derive soft-biometric information, \eg gait \cite{Delgado21}. Many works have reported impressive results, \eg in \cite{Morales20,Bortolato-FeatureDisentanglement-2020}, while the privacy protection capabilities of proposed methods are usually tested using simple machine learning-based classifiers and visualisations of dimensionality reduction tools. It has been argued that a more rigorous analysis is necessary to measure the actual privacy enhancement provided by such techniques \cite{TerhoerstBIOSIG}. Moreover, it has already been shown that different soft-biometric privacy enhancement methods can be detectable \cite{Rot2022} and vulnerable to elaborated attacks \cite{Osorio21}.

\section{Deep Learning-based attacks}\label{sec:attacks}

Biometric systems are challenged by numerous  potential attack vectors, see \cite{BRatha01a}. Some of the common attacks, \eg presentation attacks, require knowledge about the original biometric data $M$ of the target subject to be attacked. In biometric template protection schemes, $M$ is replaced by a protected biometric reference $[\mathbf{PI},\mathbf{AD}]$ such that said attacks become less relevant. However, it was recently shown that deep learning techniques can be applied to create artificial biometric samples for which the probability of achieving a false match when being compared against  unknown biometric references is significantly higher than that of a random impostor. Inspired by the concept of master keys, different researchers proposed to train a Generative Adversarial Network (GAN) to create such biometric samples. These samples can then be applied to launch dictionary attacks on biometric verification systems that allow multiple authentication attempts on biometric identification systems. The effectiveness of such attacks  has been demonstrated for fingerprints \cite{8698539} and face images \cite{9304893}. Theoretically, such attacks can be applied to template protection systems, too.

As mentioned earlier, a false match with $[\mathbf{PI},\mathbf{AD}]$ usually makes it possible for an attacker to reconstruct a close approximation of $M$. In most cases, $M$ is a feature vector of a certain type, \eg binary. This means, an attacker may additionally want to reconstruct the corresponding biometric sample, \eg fingerprint or face image. Such a sample could then be misused by the attacker to launch further attacks in particular presentation attacks \cite{Marcel-AntiSpoofing-2019}.

Developments in deep learning have shown impressive results for reconstructing biometric samples from their corresponding feature vectors. Particularly deconvolutional neural networks can be employed to reconstruct biometric samples from their corresponding embeddings in the latent space, \eg for face in \cite{Mai19a}. Further, deep learning-based methods can be employed to reconstruct biometric samples from conventional feature representations, \eg minutiae sets in \cite{Kim18}. For different biometric characteristics, it has been shown that this is even possible for binarised feature vectors. For instance, specific deep learning-based methods have been proposed to reconstruct face and vein images from binary feature vectors in \cite{Keller21} and \cite{Kauba21}, respectively. This poses a severe security risk which necessitates strong protection of biometric reference data to prevent from misuse, \eg identity fraud. In biometric template protection schemes, said reconstruction attacks become a serious threat, especially in a full disclosure attack model. In this attack model, the attacker is in possession of the secret $k$ such that the original biometric data $M$ can likely be reconstructed \cite{ISO18-TemplateProtection}. Directly reconstructing biometric samples from protected templates with such approaches may be infeasible. However, reconstruction methods can be applied after launching a false accept attack \cite{Keller21}.

The aforementioned attacks can be applied as part of an attack on biometric template protection. However, so far no deep learning-based methods that directly attack template protection schemes have been published in the scientific literature. Nevertheless, it is reasonable to assume that such attacks may soon be proposed.     

Finally, it is worth noting that the improved generalisation capabilities of deep learning-based biometric systems can lead to robustness against certain perturbations which have been introduced for the purpose of template protection. Such perturbations have frequently been proposed for cancelable biometrics \cite{Patel-CancelableBiometrics-2015}. For instance, it has been shown in \cite{Kirchgasser20b} that current deep learning-based face recognition systems are to a certain degree robust against the application of image warping which has previously been suggested to obscure and, hence, protect facial data \cite{BRatha01a}.   

\section{Further Works}\label{sec:further}
Some works in the area of deep learning are focusing on further processing steps that are vital in biometric template protection schemes. One example is pre-alignment of biometric data. Since template protection schemes protect and, therefore, conceal the original biometric data it is often impossible to perform an alignment during comparison. This is particularly the case for biometric cryptosystems. Alignment processes at comparison stage are for instance commonly required for pairwise comparisons of fingerprints \cite{Tams13} or irides \cite{Drozdowski17}. Since the original biometric data is ``unseen'' during the comparison of protected templates, pre-alignment methods are necessary.   

In \cite{Schuch18,Dieckmann19}, different deep learning techniques have been investigated for the purpose of fingerprint alignment. In particular, a fingerprint pre-alignment method which can be employed as pre-processing step  in fingerprint-based template protection systems was proposed in  \cite{Dieckmann19}. Similar approaches could be applied to other biometric characteristics.

As mentioned in the beginning, homomorphic encryption can be used to permanently protect biometric data. Homomorphic encryption makes it possible to compute basic operations in the encrypted domain which are functionally
equivalent to those in the plaintext domain and thus enables
the estimation of certain distances between protected biometric
templates. In addition, it has been shown that deep Learning-based classification tasks can be performed on homomorphically encrypted data \cite{Gilad-Bachrach16}. These so-called CryptoNets could also be applied to perform neural network-based classifications on homomorphically encrypted biometric data, \eg prediction of soft-biometric attributes.  Motivated by that fact that homomorphic encryption does not achieve irreversibility when the secret $k$ is known to an attacker, researchers have recently proposed to apply homomorphic encryption on top of cancelable biometrics \cite{OtroshiHybrid2022,BassitHybrid2022}. Thereby, some protection is retained, even in the full disclosure attack model. 

\section{Summary}\label{sec:summary}

The interest in biometric technologies has been steadily growing in recent years. Systems incorporating biometric technologies have become ubiquitous in personal, commercial, and governmental identity management applications. Deep neural networks currently represent the most popular method for pattern recognition and are applied in various biometric systems.

In response to privacy concerns raised against the (mis-)use of biometric data, template protection schemes that embody fundamental data protection principles have been introduced. This work provides an overview of how recent developments in deep learning affect biometric template protection. Several research directions are discussed in which deep learning-based techniques can be employed to improve template protection systems. In addition, deep learning-based attacks are described. By focusing on recent trends of deep learning in the research field of template protection, this survey is intended to complement existing works that review template protection methods published in the scientific literature.     

\section*{Acknoledgements}\label{sec:ack}
This research work has been funded by the German Federal Ministry of Education and Research and the Hessian Ministry of Higher Education, Research, Science and the Arts within their joint support of the National Research Center for Applied Cybersecurity ATHENE.

\bibliographystyle{IEEEtran}
\bibliography{references}


\end{document}